\documentclass{article}

    \PassOptionsToPackage{numbers, compress}{natbib}
 \usepackage[preprint]{neurips_2025}

\usepackage[utf8]{inputenc} 
\usepackage[T1]{fontenc}    
\usepackage{hyperref}       
\usepackage{url}            
\usepackage{booktabs}       
\usepackage{amsfonts}       
\usepackage{nicefrac}       
\usepackage{microtype}      
\usepackage{xcolor}         

\usepackage{amsmath}
\usepackage{bbm}
\usepackage{multirow}
\usepackage[pdftex]{graphicx}

\title{Sequential Policy Gradient for Adaptive Hyperparameter Optimization}

\author{%
  Zheng Li
    , \quad Jerry Cheng, \quad Huanying Gu \\
  Department of Computer Science\\
  New York Institute of Technology\\
  New York, NY 10023 \\
  \texttt{zli66@nyit.edu, \quad jcheng18@nyit.edu, \quad hgu03@nyit.edu} \\
}

\begin{document}

\maketitle

\begin{abstract}
  Reinforcement learning is essential for neural architecture search and hyperparameter optimization, but the conventional approaches impede widespread use due to prohibitive time and computational costs.
  Inspired by DeepSeek-V3 multi-token prediction architecture, we propose Sequential Policy Gradient modeling (SPG), a novel trajectory generation paradigm for lightweight online hyperparameter optimization.
  In contrast to conventional policy gradient methods, SPG extends the base model with temporary modules, enabling it to generate state-action (padded) trajectories in a single forward pass.
  Our experiments demonstrate that models gain performance when retrained with SPG on their original datasets and also outperform standard transfer fine-tuning. We evaluate on five datasets spanning computer vision (ImageNet, COCO), natural language processing (GLUE, SQuAD), and audio (SUPERB) to assess the industrial applicability of SPG. The proposed method demonstrates consistent improvements across widely adopted models, achieving performance gains of $+0.2\sim7\%$, with significantly low computational costs. 
  Fully reproducible code and pre-trained models:  {\color{blue}\url{https://huggingface.co/UniversalAlgorithmic/SPG}}.
\end{abstract}

\section{Itroduction}

Reinforcement learning has achieved remarkable success in Neural Architecture Search (NAS) \cite{wang2018skipnet,caiproxylessnas,mnih2014recurrent,bengio2016conditional}, as well as in Hyperparameter Optimization (HPO) for either auto data augmentation policies \cite{cubuk2019autoaugment} or learning rate adaptation \cite{daniel2016learning,bello2017neural}.
These reinforcement learning approaches significantly reduce the computational costs associated with traditional NAS and grid search methods. Although most studies have only been benchmarked on small-scale datasets \cite{lecun1998gradient,xiao2017fashion,netzer2011reading,krizhevsky2009learning}, they still represent substantial advancements.

Despite recent progress, applying reinforcement learning to NAS and HPO still presents several key challenges.
First, the computational cost of collecting episodes remains prohibitively high \cite{schulman2015trust,schulman2017proximal}, even for simple algorithms such as Williams's REINFORCE \cite{williams1992simple}, primarily due to temporal complexity.
Second, conventional approaches typically optimize hyperparameters for batched input data. Yet growing evidence suggests that optimal architectures and hyperparameters are highly input-dependent \cite{mnih2014recurrent,wang2018skipnet}. For instance, harder-to-predict inputs may require milder data augmentation, higher learning rates, and lower dropout probabilities during training.
Finally, most existing policy optimization-based methods for NAS and HPO focus narrowly on simplistic and homogeneous tasks, limiting their applicability to complex and diverse real-world scenarios.

Inspired by the Multi-token Prediction (MTP) framework of DeepSeek-V3 \cite{liu2024deepseek}, we present Sequential Policy Gradient (SPG) modeling, which integrates episode generation and termination conditions directly into the model's forward pass, significantly reducing temporal overhead and training costs.
In contrast to truncated trajectory methods \cite{schulman2015trust,schulman2017proximal} that trade completeness for efficiency, we propose an input-to-multi-sequence architecture to generate complete episodes within the computational graph.
Specifically, our model features a cross-module shared head, which we repurpose as the policy network. Each MTP module enumerates hyperparameters at growing scales, and each sequential output is reformulated as a padded trajectory.
We also reformulate the gradient of the log-probability (Grad-Log-Prob) in the REINFORCE algorithm to enable adaptive HPO: assigning distinct dropout rates at fine-grained levels (\textit{e.g.,} image-level dropout for classification, pixel-level dropout for semantic segmentation, and token-level dropout for causal language modeling).

We fundamentally redefine the reward and return function used in the existing reinforcement-learning-based NAS/HPO approaches. As a traditional understanding, comparing the prediction accuracy of subnetworks guides the architecture selection for a given input dataset--the instruction is to select the simplest network that achieves correct predictions.  However, no theoretical evidence supports this assumption. Statistical confidence in the selected architecture can only be rigorously validated if the input data exhibits consistent prediction performance across all candidate deeper architectures. Similarly, correct predictions under aggressive dropout configurations may result from random guessing unless performance remains stable across all reduced dropout rates.

We evaluate SPG on large-scale datasets. On the ImageNet \cite{russakovsky2015imagenet} dataset, ResNet-50 \cite{he2016deep} retrained with SPG achieves a +1.1\% accuracy improvement over the pre-trained baseline, while the retrained EfficientNet-V2-M \cite{tan2021efficientnetv2} reaches 85.22\% accuracy, surpassing previous state-of-the-art results. On COCO \cite{lin2014microsoft} dataset, FCN \cite{long2015fully} and DeepLab-V3 \cite{chen2017rethinking} improve by 0.5\% mIoU after five epochs of model training, with particularly significant gains (>2\% IoU) for objects exhibiting high pose variability (\textit{e.g.,} humans and cats). For transfer learning, on GLUE \cite{wang2019glue}, SQuAD \cite{rajpurkar2018know}, and SUPERB \cite{yang2021superb}, BERT \cite{devlin2019bert} and Wav2Vec2 \cite{baevski2020wav2vec} models trained via SPG consistently outperform the vanilla fine-tuning method, with an average improvement of +3\% across diverse evaluation metrics.
Our extensive experiments demonstrate that SPG achieves robust performance, training efficiency, and industrial applicability.

\begin{figure}[hbtp]
  \centering
  \includegraphics[width=0.9\textwidth]{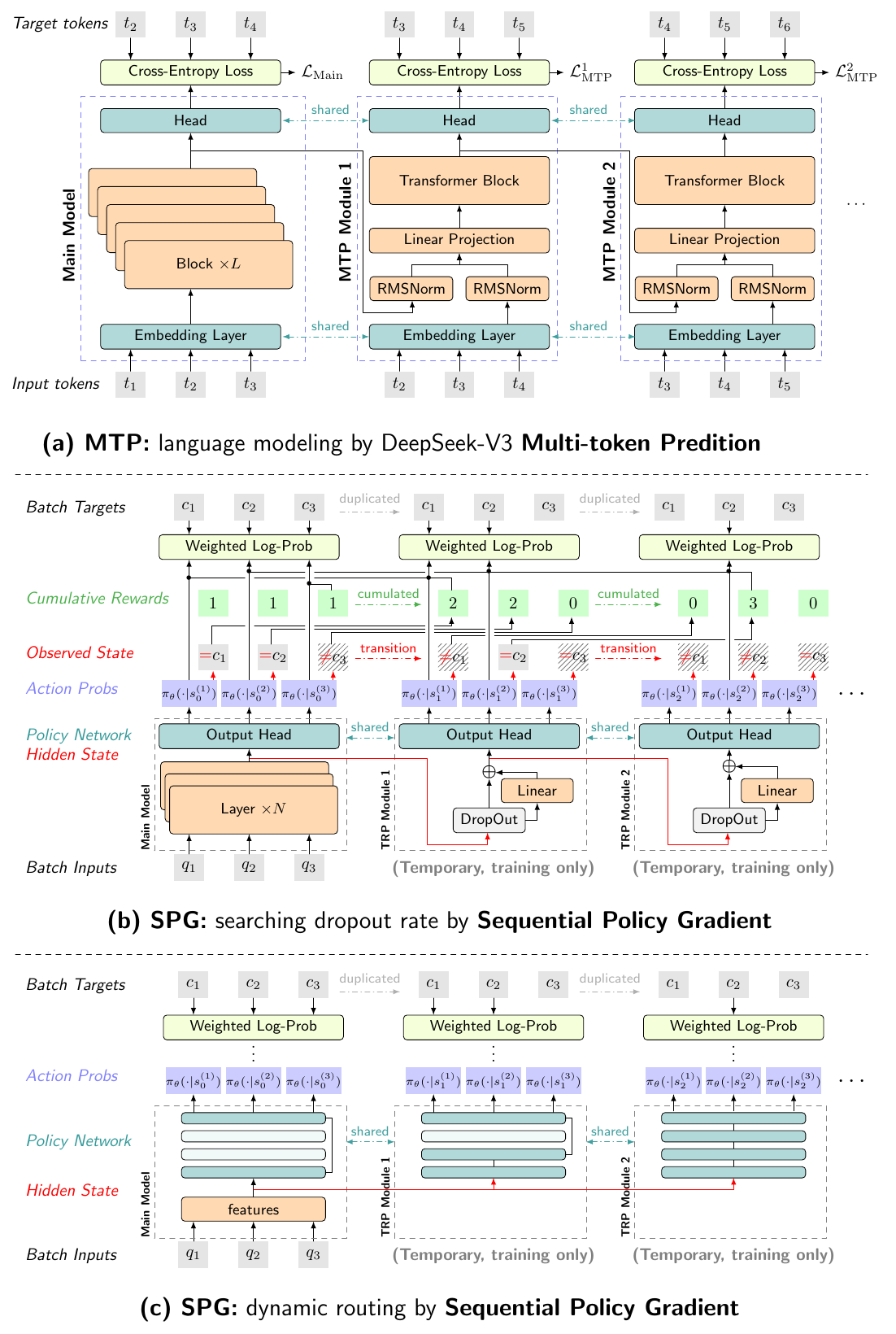}
  \caption{Illustration of the DeepSeek-V3 Multi-Token Prediction (MTP) and our Sequential Policy Gradient (SPG) implementation. 
  (a) MTP applied to causal language modeling: MTP extends the prediction scope to multiple future tokens at each position.
  (b) SPG applied to HPO: the Task Replica Prediction (TRP) modules enumerate hyperparameter values in increasing order from left to right, then sequentially produce predictions corresponding to each hyperparameter configuration. This framework generates padded episodes within a single model's forward pass.
  (c) SPG applied to NAS: the state representation and reward calculation process are identical to those in (b), without requiring repeated draws. TRP modules enumerate the model depth in increasing order from left to right.}
  \label{fig:sequential-policy-gradient}
\end{figure}

\section{Related work}

\textbf{Neural Architecture Search} is an automated machine learning (AutoML) technique \cite{zoph2022neural,liu2018darts,pham2018efficient,feurer2015efficient,snoek2012practical} designed to algorithmically discover neural network architectures (\textit{e.g.,} layer depth, kernel size, layer connection patterns), replacing manual architecture design. Its primary objective is to identify optimal architectures for recognition tasks.
NAS approaches fundamentally differ from traditional ensemble methods \cite{breiman1996bagging,freund1997decision,breiman2001random,friedman2001greedy,chen2016xgboost}. Instead of naively combining multiple networks (which incurs prohibitive computational costs, \textit{e.g.,} thousands of GPU hours), they focus on localized architecture selection.
These methods employ a gating module to regulate cross-layer connections \cite{liu2018darts,wang2018skipnet,wu2018blockdrop} or perform module selection in multi-branch architectures \cite{shazeer2017outrageously,caiproxylessnas}, where the gating layer/network is typically learned in two modes: by vanilla backpropagation \cite{liu2018darts,shazeer2017outrageously} or by policy optimization \cite{caiproxylessnas,wang2018skipnet,wu2018blockdrop}. NAS has successfully inspired the design of large language models \cite{du2022glam,achiam2023gpt,liu2024deepseek}.

\textbf{Hyperparameter Optimization} aims to automate the hyperparameter tuning process, reducing the time that developers spend on tuning hyperparameters while improving model performance \cite{hutter2019automated}. In contrast to early approaches like Grid Search \cite{hsu2003practical} and Random Search \cite{bergstra2012random}, most current HPO methods employ a proxy model to output hyperparameters \cite{snoek2012practical,hutter2011sequential,li2018hyperband,falkner2018bohb}. Inspired by this, some researchers perform the proxy model as a policy network and optimize it via reinforcement learning algorithms \cite{daniel2016learning,bello2017neural}.

\textbf{Efficient Policy Network Training.} The training of policy networks typically incurs high time costs. In contrast to training deep learning models, which only require epoch-loop iterations, reinforcement learning introduces an additional episode-loop overhead. Two predominant methodologies aim to resolve this issue: (1) Truncated Trajectories \cite{schulman2015trust,schulman2017proximal} that impose a fixed-length segment per episode, avoiding inefficient long-horizon interactions while maintaining valid policy gradient estimates; (2) Parallel Implementations to reduce the total time costs \cite{mnih2016asynchronous,espeholt2018impala}. These techniques have become foundational parts of modern reinforcement learning, balancing efficiency and performance.

\textbf{Multi-token Prediction} is a framework that enables models to predict multiple future tokens simultaneously \cite{gloeckle2024better}. While traditional language models typically generate tokens autoregressively one by one, MTP allows the model to output multiple consecutive tokens in a single forward pass, improving training efficiency and generation quality. In this framework, the output layer is extended to multiple heads, making the model input-to-multi-sequence. The authors of DeepSeek-V3 \cite{liu2024deepseek} upgraded this technique by connecting sequentially multiple extra modules with the base model, where each module shares the same head and takes the embeddings from the previous module as input. Technical report of DeepSeek-V3 shows that this powerful model requires significantly fewer GPU hours for training, yet achieves comparable performance to state-of-the-art large language models such as Qwen2.5 \cite{bai2023qwen}, GPT-4o \cite{achiam2023gpt}, and Llama-3.1 \cite{meta2024llama3}.

\section{Method}

In this section, we introduce our efficient reinforcement learning algorithm, the Sequential Policy Gradient.

\subsection{Reinforcement learning}

Fig. {\color{red}\ref{fig:sequential-policy-gradient} b} illustrates our SPG method for assigning distinct dropout rates to different inputs. The architecture consists of a base model and multiple temporary Task Replica Prediction (TRP) modules. Each TRP module corresponds to a fixed hyperparameter configuration applied to searching. All TRP modules will be removed after training.
Our primary goal is to \textit{adaptively optimize a hyperparameter for varying input on a fine-grained level}.




\textbf{TRP Modules}. To specifically introduce the hyperparameter optimization of searching the dropout rate, our TRP implementation uses $T$ temporary modules to do $T$ additional prediction task replicas, where the $t$-th TRP module consists of a shared
output head $\text{OutHead}(\cdot)$, a dropout layer $\text{Dropout}(\cdot; p_t)$ with rate $p_t$, and a linear layer $\text{Linear}(\cdot)$. For the $i$-th input $q_i$, we denote the representation of $q_i$ at the $(t-1)$-th replica by $\boldsymbol{h}_{t-1}^{(i)}$, the randomly omitted hidden state is
\begin{equation}
    \boldsymbol{h'}_t^{(i)} = \text{Dropout}(\boldsymbol{h}_{t-1}^{(i)}; p_t).
    \label{eqn:1}
\end{equation}
Especially, when $t = 1$, $\boldsymbol{h}_{t-1}^{(i)}$ refers to the representation given by the main model. Subsequently, we combine $\boldsymbol{h'}_ t^{(i)}$ with a dimension-preserving linear layer, and serve the result as the representation at the $t$-th replica:
\begin{equation}
    \boldsymbol{h}_t^{(i)} = \boldsymbol{h'}_t^{(i)} + \text{Linear}(\boldsymbol{h'}_t^{(i)}),
    \label{eqn:2}
\end{equation}
where the trainable weights and bias of the linear layer $\text{Linear}(\cdot)$ are initialized with zero values:
\begin{equation}
    [\boldsymbol{h}_t^{(i)}]_{\text{init}} = [\boldsymbol{h'}_t^{(i)}]_{\text{init}} + \text{Linear}_{\text{init}}([\boldsymbol{h'}_t^{(i)}]_{\text{init}}) = [\boldsymbol{h'}_t^{(i)}]_{\text{init}} = \text{Dropout}([\boldsymbol{h}_{t-1}^{(i)}]_{\text{init}}; p_t).
\end{equation}
Our multi-scale dropout rates are cumulative from lower to higher. For each TRP module, the input representation serves as the randomly omitted outputs of the previous module. Thus, for the $t$-th module, the input representation has been processed by $t-1$ dropout layers:
\begin{align}
    [\boldsymbol{h}_t^{(i)}]_{\text{init}} 
    &= \text{Dropout}([\boldsymbol{h}_{t-1}^{(i)}]_{\text{init}}; p_t) \nonumber \\
    &= \text{Dropout}(\text{Dropout}([\boldsymbol{h}_{k-2}^{(i)}]_{\text{init}}; p_{t-1}); p_t) \nonumber \\
    &= \text{Dropout}([\boldsymbol{h}_{k-2}^{(i)}]_{\text{init}}; 1-(1-p_{t-1}) (1-p_t)) \nonumber \\
    &= \cdots \notag \\
    &= \text{Dropout}([\boldsymbol{h}_{0}^{(i)}]_{\text{init}}; 1 - \Pi_{k=1}^{t}(1-p_k)).
\end{align}
Finally, we compute the probability distribution for the $t$-th prediction task replica by $\pi_t^{(i)}$:
\begin{equation}
    \pi_t^{(i)} = \text{OutHead}(\boldsymbol{h}_t^{(i)}) \in \mathbb{R}^{V},
    \label{eqn:5}
\end{equation}
where $V$ denotes the dimension of the output space (\textit{e.g.,} the vocabulary size, the number of image classes, or the number of instance classes, etc.), depending on the task.

\textbf{State and Action}. For the $i$-th input $q_i$, a complete state $\boldsymbol{s}_t^{(i)}=(\boldsymbol{h}_t^{(i)}, \boldsymbol{o}_t^{(i)})$ consist of the hidden state $\boldsymbol{h}_t^{(i)}$ (we represent the state transition by equations ({\color{red}\ref{eqn:1}}) and ({\color{red}\ref{eqn:2}})) and a ternary observed state $\boldsymbol{o}_t^{(i)}$ (initialize $\boldsymbol{o}_0^{(i)}$ to 1). 

In SPG modeling, if an episode terminates early, the trajectory will be padded to a fixed length $T+1$  to ensure consistent batch processing. Base on this mechanism, we define the ternary observed state to control the termination condition of the episode: when $\boldsymbol{o}=1$, allowing the trajectory to continue; when $\boldsymbol{o}=0$, terminate the trajectory normally, and when $\boldsymbol{o}=-1$, it means the episode is over, but the model still force infers the next state-action pair. The transition of this observed state can then be expressed by
\begin{equation}
    \boldsymbol{o}_{t+1}^{(i)} = \boldsymbol{o}_t^{(i)} \cdot (\boldsymbol{o}_t^{(i)} + \mathbbm{1}_{\text{arg~max~}\pi_t^{(i)} = c_i}) - 1,
    \label{eqn:6}
\end{equation}
where $\text{arg~max~}\pi_t^{(i)}$ is the prediction of the input $q_i$ on the $t$-th TRP module.
However, this prediction is confident if and only if all preceding modules (from base model to $t-1$ TRP module) simultaneously yield correct predictions, since each prediction is computed using the previous module's hidden state output.
For instance, if $s_{t+1}^{(i)}$ is the terminal state, any dropout rate greater than $1 - \Pi_{k=1}^{t} (1 - p_k)$ should be rejected for model training, because the prediction of input $q_i$ has low confidence during the current training epoch -- even if the prediction is correct, it may merely result from random guessing.

Subsequently, we define the action space as 
\begin{equation*}
    \mathcal{A} = \{ \text{continue to retrieve the hyperparameters of the next scale}, \text{stop retrieving} \},
\end{equation*}
and the policy for input-target pair $(q_i, c_i)$ with parameters $\theta$ is expressed by
\begin{equation}
    a_t^{(i)} \sim \pi_{\theta} (\cdot | s_t^{(i)}) = ([\pi_t^{(i)}]_{c_i}, 1 - [\pi_t^{(i)}]_{c_i}).
    \label{eqn:7}
\end{equation}

\textbf{Trajectories.} Our architecture infers multiple parallel trajectories in a single forward pass. For the $i$-th input $q_i$, the analytical form of state transition $f: (s_t^{(i)}, a_t^{(i)}) \mapsto s_{t+1}^{(i)}$ is given by the combination of equations ({\color{red}\ref{eqn:1}-\ref{eqn:2}}) and ({\color{red}\ref{eqn:5}-\ref{eqn:7}}).

Trajectories are independent of each other, and the step length of each whole episode can be different. As depicted on the bottom side of Fig. {\color{red}\ref{fig:sequential-policy-gradient}}, we have three parallel trajectories $\tau^{(1)} = (s_0^{(1)}, a_0^{(1)}, s_1^{(1)}, a_1^{(1)}, s_2^{(1)})$, $\tau^{(2)} = (s_0^{(2)}, a_0^{(2)}, s_1^{(2)}, a_1^{(2)}, s_2^{(2)}, a_2^{(2)}, s_3^{(2)})$, and $\tau^{(3)} = (s_0^{(3)}, a_0^{(3)}, s_1^{(3)})$. 

\textbf{Reward and Return}. In conventional categorical policies, failure results in a very negative reward for the entire episode, which can significantly increase the gradient variance. We define a relatively flat reward and reformulate the undiscounted return. 

Since every whole episode is padded to a fixed length $T$ (as previously described), we set the reward of each dummy state to zero:
\begin{equation}
    r_t^{(i)} = R(s_t^{(i)}) = \mathbbm{1}_{\boldsymbol{o}_t^{(i)} \ge 0}.
    \label{eqn:8}
\end{equation}
We represent the return by the (weighted) sum of the binary rewards. For a padded $t$-step trajectory $\tau^{(i)} = (s_0^{(i)}, a_0^{(i)}, \cdots, s_{t}^{(i)})$, the undiscounted return is also reset to zero when it ends with a dummy state:
\begin{equation}
    R(\tau^{(i)}) = (r_{0}^{(i)} + \cdots + r_{t-1}^{(i)}) r_{t-1}^{(i)}.
    \label{eqn:9}
\end{equation}

\textbf{Grad-Log-Prob of Group Trajectories.} Given a trajectory set $\mathcal{D}$, the policy gradient is estimated as:
\begin{equation*}
    \nabla_{\theta} \mathcal{J}(\pi_{\theta}) \approx \frac{1}{|\mathcal{D}|} \sum_{\tau \in \mathcal{D}} \nabla_{\theta} \text{log~} P(\tau|\theta) = \frac{1}{|\mathcal{D}|} \sum_{\tau \in \mathcal{D}} \sum_{t=0}^{|\tau|-1} \nabla_{\theta} \text{log~} \pi_{\theta}(a_t, s_t) R(\tau).
\end{equation*}
where $\nabla_{\theta} \text{log~} P(\tau|\theta)$ is the Grad-Log-Prob of a trajectory $\tau$, ${|\tau|}$ is the step length of $\tau$, and $|\mathcal{D}|$ is the number of trajectories in $\mathcal{D}$.

When employing Monte Carlo rollouts to collect trajectories, the gradient estimate becomes more accurate due to the small magnitude of $|\mathcal{D}|$. From equations ({\color{red}\ref{eqn:6}}), ({\color{red}\ref{eqn:8}}), and ({\color{red}\ref{eqn:9}}), it follows that:
\begin{equation}
    \{ \tau \in \mathcal{D}~|~R(\tau) \neq 0 \} = \{ \tau \in \mathcal{D}~|~a_0 = \cdots = a_{|\tau|-3}, \boldsymbol{o}_{|\tau|-1} \ge 0 \}.
    \label{eqn:10}
\end{equation}
For a batch of inputs $\{q_i\}_{i=1}^{m}$, our architecture parallelly generates a group of independent trajectories $\{ \tau^{(i)} \}_{i=1}^{m}$ in a single forward propagation, diverging from the online training mode.
In this paper, ``batch size'' refers to the total number of images/pixels/tokens processed per update, depending on the recognition task (see Sec. {\color{red}\ref{sec:implementation}} for details).
Here, we compute the mean of the Grad-Log-Prob from batch trajectories to reduce the variance of the total gradient:
\begin{equation}
    \overline{\nabla_{\theta} \text{log~} P(\tau|\theta)} \overset{\Delta}{=} \frac{1}{m} \sum_{i=1}^{m} \frac{1}{|\tau^{(i)}|} \nabla_{\theta} \text{log~} P(\tau^{(i)}|\theta).
\end{equation}

We do not explicitly present our reinforcement learning algorithm, as it is a reformulation of Williams's REINFORCE algorithm \cite{williams1992simple} with Grad-Log-Prob replaced by $\overline{\nabla_{\theta} \text{log~} P(\tau|\theta)}$.

\begin{figure}[ht]
  \centering
  \includegraphics[width=1.0\textwidth]{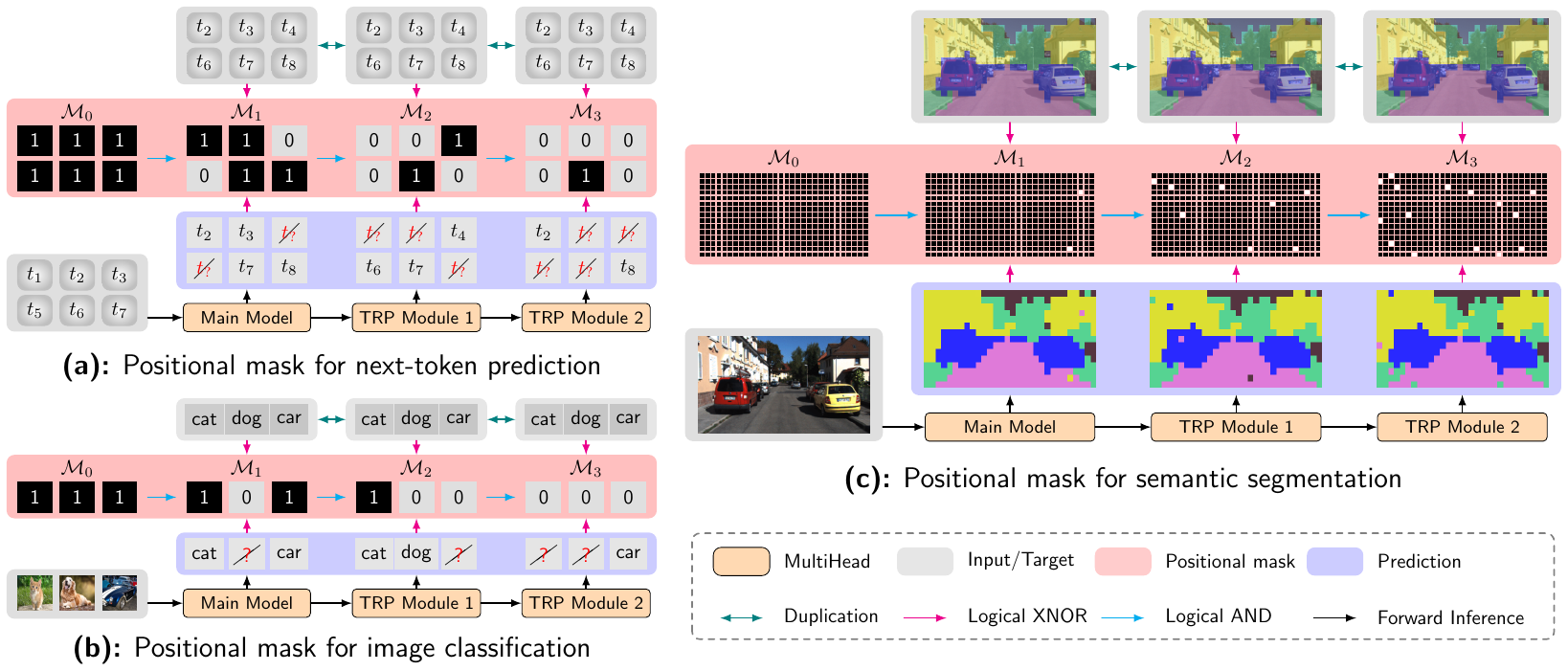}
  \caption{\textbf{Positional Mask.} (a) in causal language modeling: positional masks are 2-by-3 matrices, corresponding to a batch size of 2 and the sequence length of 3; (b) in image classification: positional masks are 3-dimensional vectors, matching the batch size of 3; in causal semantic segmentation: positional masks are tensors of shape [1, 180, 360, 2], corresponding to a batch size of 1, an image height of 180 pixels, a width of 360 pixels, and dual prediction outputs.}
  \label{fig:positional-mask}
\end{figure}

\subsection{Implementation}
\label{sec:implementation}

\textbf{Positional Mask.} As aforementioned, the reward in our SPG algorithm is computed on a fine-grained level, and the effective batch size $m$ is defined as the total number of independent units processed in parallel. As shown in Fig. {\color{red}\ref{fig:positional-mask}}, for image classification, the effective batch size $m$ is simply the number of images per batch; for semantic segmentation, the effective batch size $m = N \times H \times W \times 2$, where $N$ is the number of images, $H \times W$ is the spatial resolution, and the final factor (value 2) corresponds to the number of model outputs (standard and auxiliary); in language modeling, the effective batch size $m$ equals the number of tokens (batch size $\times$ sequence length). In this section, we define a tensor $\mathcal{M}_t \in \mathbb{R}^m$ to represent the batch processing of rewards:
\begin{equation}
    [ \mathcal{M}_t ]_i \overset{\Delta}{=} r_t^{(i)}.
\end{equation}

Fig. {\color{red}\ref{fig:positional-mask}} illustrates the pipeline of computing $\mathcal{M}$. $\mathcal{M}$ is initially filled with the scalar value 1, and then updated according to equations ({\color{red}\ref{eqn:6}}) and ({\color{red}\ref{eqn:8}}):
\begin{equation}
    [\mathcal{M}_{t+1}]_i = [\mathcal{M}_t]_i \cdot \mathbbm{1}_{\text{arg~max~}\pi_t^{(i)} = c_i}.
\end{equation}

\textbf{Temporary parameters.} As previously mentioned, all TRP modules are temporary,  exclusively used during training, and removed afterward. For vision models, we typically set the shared head as the final layer of the network; for language models, we employ either the decoder or the head as the shared head. 
In Fig. {\color{red}\ref{fig:sequential-policy-gradient} b}, the number of additional trainable parameters increases monotonically with the representation/hidden state dimension $D$, and is proportional to the number of TRP modules $T$. The extra parameters are very limited. For any base model with $N_{\text{base}}$ parameters, the total parameter count $N_{\text{total}}$ is given by:
\begin{equation}
N_{\text{total}} = N_{\text{base}} + T \cdot (D^2 + D).
\end{equation}

\section{Empirical results}
\label{sec:empirical-results}

Our experiments cover a broad range of tasks, including image classification, semantic segmentation, natural language understanding, question answering, and audio classification. In Sec. {\color{red}\ref{sec:model-retraining}}, we evaluate MobileNet-V2 \cite{sandler2018mobilenetv2}, ResNet \cite{he2016deep}, EfficientNet-V2 \cite{tan2021efficientnetv2}, and ViT \cite{dosovitskiy2020image} on ImageNet-1K \cite{russakovsky2015imagenet} dataset, FCN \cite{long2015fully} and DeepLab-V3 \cite{chen2017rethinking} on COCO 2017 \cite{lin2014microsoft} benchmarks. We follow TorchVision's settings \cite{torchvision2016} for these pre-trained models, and retrain these models via SPG, then conduct direct comparisons with the original pre-trained versions.
In Sec. {\color{red}\ref{sec:transfer-learning}}, we further demonstrate the application scope of SPG on transfer learning. Specifically, we conduct experiments using BERT \cite{devlin2019bert} on the GLUE benchmark \cite{wang2019glue} and the SQuAD dataset \cite{rajpurkar2018know}, and Wav2Vec2 \cite{baevski2020wav2vec} on the SUPERB benchmark \cite{yang2021superb}. Our results are compared against vanilla fine-tuning baselines provided by HuggingFace \cite{wolf-etal-2020-transformers}.
We also introduce the application of SPG for NAS in Sec. {\color{red}\ref{sec:neural-architecture-search}}: ResNet-18 will probabilistically evolve into ResNet-50 or ResNet-101 during training through connecting with temporary layers.
All pre-trained models are available in TorchVision and HuggingFace, and we follow similar training recipes to these pre-trained models. All experiments are conducted on NVIDIA RTX A6000 GPUs.

\subsection{Model retraining}
\label{sec:model-retraining}

\textbf{Baselines.} We adopt the pre-trained models from TorchVision as baselines and reuse their training recipes whenever possible for our SPG retraining. In the official report, the pre-trained ResNet and ViT models were trained on 64 Tesla V100 GPUs, while MobileNet-V2 and EfficientNet-V2 were trained on 8 Tesla V100 GPUs. To evaluate training costs, we convert the prior GPU hours into RTX A6000 GPU-hours based on their performance specifications.
In addition, there are different recipes in the TorchVision reports, resulting in multiple weight versions of the pre-trained MobileNet-V2, ResNet, and ViT models with varying evaluation metric values. To ensure fair ablation studies, we consider reuse recipes that apply standard techniques, such as learning rate scheduling \cite{loshchilov2022sgdr}, warmup, $L^2$ regularization, data augmentation \cite{cubuk2020randaugment,zhang2018mixup,yun2019cutmix}, and model exponential moving average \cite{izmailov2018averaging}.
Finally, for semantic segmentation, we report mean IoU and pixel-wise accuracy, where the ``background'' class is included in the metric computation.

\textbf{Setup.} All SPG models are trained with the three TRP modules: drop layers with rates $p_1 = 0.2$, $p_2 = 1 - (1 - 0.2)^2 = 0.36$, and $p_3 = 1 - (1 - 0.2)^3 = 0.488$; we applied a weighted return $R(\tau^{(i)}) = (r_{0}^{(i)} + \lambda_1 r_{1}^{(i)} + \cdots + \lambda_{t-1} r_{t-1}^{(i)}) r_{t-1}^{(i)}$ to the REINFORCE algorithm, where $\lambda_1 = 0.4$, $\lambda_2 = 0.2$, $\lambda_3 = 0.1$.
For retraining MobileNet-V2 and ResNet, the learning rate is set to 4E-4 with a maximum of 40 training epochs; for EfficientNet-V2, and ViT, the learning rate is set to 5E-9, with a maximum of 5 epochs for EfficientNet-V2 and 10 epochs for ViT; for FCN, and DeepLab-V3, the learning rate is set to 2E-3, with a maximum of five epochs. Although EfficientNet-V2 and ViT are trained with extremely small learning rates, directly fine-tuning them still leads to catastrophic forgetting. As an example, the issue arises because the optimizer for training ViT, AdamW \cite{loshchilov2017decoupled}, initializes both the first moment vector $\vec{m}$ and the second moment vector $\vec{v}$ to zero values. To address this issue, we introduce a cold-start phase with a zero learning rate before the retraining begins.

\begin{table}[!hb]
    \caption{Model comparison on the ImageNet-1K dataset. ``\# Parameters (Static)'' refers to the number of parameters used for model evaluation and inference; ``\# Parameters ({\color{gray}Temporary})'' refers to the number of extra parameters used for model training. Temporary parameters reside within TRP modules, which are removed after model training. Metrics include Top-1 validation accuracy (Acc@1) and Top-5 validation accuracy (Acc@5). Throughput denotes the number of images processed per second during training.  ``$\downarrow$'' or ``$\uparrow$'' indicates that lower or higher values are better.}
   \label{tab:imagenet}
    \centering
    \resizebox{\textwidth}{!}{
    \begin{tabular}{l|l|l|l|l|l}
    \toprule
    & \# Parameters & & & Throughput & Training Costs \\ 
    Model \& Type & (Static{\color{gray}/Temporary}) & Acc@1 (\%) & Acc@5 (\%) & (img/s)$\uparrow$ & (GPU Hours)$\downarrow$ \\
    \midrule
    MobileNet-V2$_{\text{base}}$ & 3.5 M & 71.878  & 90.286  & 901.51 & 197 \\ 
    MobileNet-V2$_{\text{SPG}}$ & 3.5 M {\color{gray}/ 4.9 M} & $72.104_{\pm 0.02}$  & $90.316_{\pm 0.07}$  & 882.12 & 26.6 \\
    \midrule
    ResNet-50$_{\text{base}}$ & 25.6 M & 76.130  & 92.862  &  440.18 & 82.8 \\ 
    ResNet-50$_{\text{SPG}}$ & 25.6 M {\color{gray}/ 12.6 M} & $77.234_{\pm 0.27}$  & $93.322_{\pm 0.03}$  & 411.50 & 36.8 \\
    \midrule
    EfficientNet-V2-M$_{\text{base}}$ & 54.1 M      & 85.112  & 97.156  & 85.73 & 4.3E+3 \\ 
    EfficientNet-V2-M$_{\text{SPG}}$ & 54.1 M {\color{gray}/ 4.9 M} & $\textbf{85.218}_{\pm (< 0.01)}$  & $\textbf{97.208}_{\pm (< 0.01)}$  & 82.95 & 71.6 \\
    \midrule
    ViT-B16$_{\text{base}}$ & 86.6 M      & 81.072  & 95.318  & 431.02 & 306 \\ 
    ViT-B16$_{\text{SPG}}$ & 86.6 M {\color{gray}/ 1.8 M} & $81.092_{\pm (< 0.01)}$  & $95.304_{\pm (< 0.01)}$  & 416.31 & 5.10 \\
    \bottomrule
    \end{tabular}
    }
\end{table}

\textbf{Cold start the optimizer.} We propose a cold-start initialization for the optimizer with moment vectors \cite{kingma2014adam,loshchilov2017decoupled}. Prior to actual training, we set the learning rate to zero and run the model for several epochs. During this phase, model parameters remain unchanged (ignoring minor perturbations such as those from BatchNorm \cite{ioffe2015batch}), while moment vectors are updated by the gradients. After the moment vectors converge, we reset the learning rate for model training to effectively mitigate catastrophic forgetting of models caused by the optimizer in the early stage of training. In our experiments, we apply this strategy only to the EfficientNet-V2, ViT, FCN, and DeepLab-V3, each for three epochs.

\textbf{Compared with pre-trained models.} SPG improves model performance with limited training costs and is orthogonal to existing techniques such as data augmentation, knowledge distillation, and model ensemble.
This section presents a concise experimental report:
(1) On the 1000-categories classification task of ImageNet-1K, SPG improves the Top-1 accuracy of ResNet-50 by $1.1\%$, as shown in Tab. {\color{red}\ref{tab:imagenet}};
(2) For EfficientNet-V2, SPG further updates the Top-1 accuracy to $85.218\%$ with only five training epochs, where the first three epochs serve as a cold-start phase;
(3) We also evaluate semantic segmentation models on a subset of COCO val2017, on the 20 categories that are present in the Pascal VOC dataset \cite{everingham2010pascal}. The results are summarized in Tab. {\color{red}\ref{tab:coco}};
(4) Due to the specific design of linear layers in TRP, SPG introduces only a small number of temporary parameters when applied to retrain the ViT, resulting in a relatively modest performance gain compared to other models.

\begin{table}[!t]
    \caption{Training from Scratch vs. SPG Retraining: performance comparison on semantic segmentation models. We conduct evaluations on a subset of COCO val2017, on the 20 categories that are present in the Pascal VOC dataset. SPG improves model performance compared with the baselines, particularly for objects with highly variable poses (\textit{e.g.,} person and cat), where the IoU gain exceeds 2\%.}
   \label{tab:coco}
    \centering
    \resizebox{0.9\textwidth}{!}{
    \begin{tabular}{l|l|l|l|l|l}
    \toprule
    & & \# Parameters & mean & pixelwise & Training Costs \\ 
    Model & Type & (Static{\color{gray}/Temporary}) & IoU (\%) & Acc (\%) & (GPU Hours)$\downarrow$ \\
    \midrule
    FCN-ResNet50 & base & 35.3 M & 60.5 & 91.4 & 63.84 \\
    FCN-ResNet50 & SPG & 35.3 M {\color{gray}/ 15.7 M} & $60.9_{\pm 0.3}$ & $91.6_{\pm 0.1}$ & 10.64 \\
    FCN-ResNet101 & base & 54.3 M & 63.7 & 91.9 & 81.20 \\
    FCN-ResNet101 & SPG & 54.3 M {\color{gray}/ 15.7 M} & $64.3_{\pm 0.1}$ & $91.9_{\pm (< 0.1)}$ & 13.53 \\
    \midrule
    DeepLabV3-ResNet50 & base & 42.0 M & 66.4 & 92.4 & 87.28 \\
    DeepLabV3-ResNet50 & SPG & 42.0 M {\color{gray}/ 15.7 M} & $66.6_{\pm 0.2}$ & $92.5_{\pm 0.1}$ & 14.53 \\
    DeepLabV3-ResNet101 & base & 61.0 M & 67.4 & 92.4 & 103.9 \\
    DeepLabV3-ResNet101 & SPG & 61.0 M {\color{gray}/ 15.7 M} & $\textbf{67.8}_{\pm 0.1}$ & $\textbf{92.5}_{\pm (< 0.1)}$ & 17.33 \\
    \bottomrule
    \end{tabular}
    }
\end{table}

\subsection{Transfer learning}
\label{sec:transfer-learning}

\begin{table}[!hb]
    \caption{Performance of models for transfer learning trained with fine-tuning (FT) \textit{vs.} SPG.
    The top 12 rows compare the performance of the BERT-base on various language understanding tasks (\textit{e.g.,} CoLA, SST-2, etc.) from the GLUE benchmark, using FT and our proposed SPG method.
    Entries marked with $*$ denote results of the BERT-base on the SQuAD-1.0 dataset for question answering (Q/A).
    Entries marked with $\dagger$ indicate results of the Wav2Vec2-base on the keyword spotting subset of the SUPERB dataset for audio classification (AC).}
   \label{tab:glue}
    \centering
    \resizebox{0.7\textwidth}{!}{
    \begin{tabular}{l|l|l|l|l}
    \toprule
    & & \multicolumn{2}{c|}{} & GPU Hours \\ 
    Task & Metric Type & \multicolumn{1}{c}{FT (\%)} & \multicolumn{1}{c|}{SPG (\%)} & (FT{\color{gray}/SPG})$\downarrow$ \\
    \midrule
    CoLA & Matthews coor & 56.53 & 62.13 & 0.06 {\color{gray}/ 0.12} \\
    \midrule
    SST-2 & Accuracy & 92.32 & 92.54 & 0.43 {\color{gray}/ 0.72} \\
    \midrule
    MRPC & F1/Accuracy & 88.85/84.09 & 91.10/87.25 & 0.04 {\color{gray}/ 0.05} \\
    \midrule
    QQP & F1/Accuracy & 87.49/90.71 & 89.72/90.88 & 2.36 {\color{gray}/ 7.86} \\
    \midrule
    QNLI & Accuracy & 90.66 & 91.10 & 0.67 {\color{gray}/ 2.22} \\
    \midrule
    RTE & Accuracy & 65.70 & 72.56 & 0.02 {\color{gray}/ 0.05} \\
    \midrule
    Q/A$^*$ & F1/Extra match & 88.52/81.22 & 88.67/81.51 & 0.91 {\color{gray}/ 0.90} \\
    \midrule
    AC$^{\dagger}$ & Accuracy & 98.26 & 98.31 & 0.23 {\color{gray}/ 0.21} \\
    \bottomrule
    \end{tabular}
    }
\end{table}

\textbf{Baselines.} We use the pre-trained BERT-base (110M parameters) and Wav2Vec2-base (95M parameters) models from the HuggingFace Transformers library, follow the random seed (42) and training recipes provided in the HuggingFace examples. According to their documentation, the fine-tuning of pre-trained models is performed on a single GPU, we adopt the same setup in experiments.

\textbf{Compared with vanilla fine-tuning.}
We adopt the SPG configuration described in Sec. {\color{red}\ref{sec:model-retraining}}. Specifically, we set the learning rates as follows: 2.5E-5 for CoLA, 3E-5 for SST-2, 2E-5 for MRPC, 1E-5 for QQP, 2E-5 for QNLI, and 5E-5 for RTE. Since the embedding dimension of BERT-base is 768, the number of temporary parameters introduced by SPG is calculated as $768 \times 769 \times 3 \approx 1.8$ M, which also applies to Wav2Vec2-base as well. The proposed SPG demonstrates consistent superiority over conventional fine-tuning across diverse transfer learning tasks, with particularly pronounced gains on simpler tasks, as shown in Table {\color{red}\ref{tab:glue}}. This suggests that the SPG's margin of improvement inversely correlates with task complexity, implying stronger robustness in low-complexity settings.

\subsection{Neural architecture search}
\label{sec:neural-architecture-search}

We investigate the feasibility of applying SPG to NAS. As shown in Fig. {\color{red}\ref{fig:sequential-policy-gradient}}c, with ResNet as the base model, we sequentially append three temporary TRP modules, both kernel weights and bias of convolution layers in TRP modules are initialized with zero value. We adopt the same weighted return from Sec. {\color{red}\ref{sec:model-retraining}} and a one-epoch cold start. Specifically, we start with a learning rate of 4E-4 and divide by 0.5 at each two epochs, and terminate training at 10 epochs. The results in Table {\color{red}\ref{tab:nas}} show that with SPG retraining, ResNet-18, ResNet-34, and ResNet-50 gain extra Top-1 and Top-5 validation accuracy on ImageNet-1K, establishing new state-of-the-art results.

\begin{table}[ht]
    \caption{Performance of pre-trained \textit{vs.} SPG-retrained models on ImageNet-1K.
    Each base model is expanded through three sequential TRP modules: ResNet-18 grows to ResNet-27, then ResNet-36, and finally ResNet-45; ResNet-34 expands to ResNet-40, ResNet-46, and ResNet-52; while ResNet-50 develops into ResNet-53, ResNet-56, and ResNet-59 through three sequential module additions. After SPG retraining, we keep only the base architecture (ResNet-18, ResNet-34, and ResNet-50) and remove all expanded variants.}
   \label{tab:nas}
    \centering
    \resizebox{\textwidth}{!}{
    \begin{tabular}{l|l|l|l|l|l|l}
    \toprule
    & & & \multicolumn{3}{c|}{Cumulative Depths} &  \\ 
    Base Model & \# Parameters & Acc@1/Acc@5 (\%)  & \multicolumn{1}{c}{TRP$_1$} & \multicolumn{1}{c}{TRP$_2$} & \multicolumn{1}{c|}{TRP$_3$} & Acc@1/Acc@5 (\%) \\
    \midrule
    ResNet-18 & 11.7 M & 69.758 / 89.078 & 27 & 36 & 45 & 70.092 / 89.314 \\
    \midrule
    ResNet-34 & 21.8 M & 73.314 / 91.420 & 40 & 46 & 52 & 73.900 / 93.536 \\
    \midrule
    ResNet-50 & 25.6 M & 76.130 / 92.862 & 53 & 56 & 59 & 77.234 / 93.322 \\
    \bottomrule
    \end{tabular}
    }
\end{table}

In contrast to NAS, our method eliminates the gating network and restricts the TRP module to training phases only.
Contrary to the conventional wisdom that ``simple images require fewer layers whereas complex ones demand deeper architectures,'' we argue that ``an image correctly predicted by a shallow network does not imply that the image is simple; its predictions should remain valid (or improve) in deeper networks to avoid random-guessing behavior''.
Note that SPG fundamentally differs from knowledge distillation: the deeper ResNets in TRP modules are not teachers, as they are not pre-trained, and initialized with zero-value parameters.

\section{Limitations and future work}
\label{sec:limitations-and-future-work}
\textbf{Versatility.} This study primarily focuses on policy gradient algorithms to adaptively optimize the hyperparameter (dropout rate) and enable the neural architecture search. A potential direction for extending the application scope of SPG involves reformulating the TRP module, thereby enabling adaptive optimization of other hyperparameters (\textit{e.g.,} the negative slope $\alpha$ in Leaky ReLU \cite{maas2013rectifier} and the momentum in Batch Normalization \cite{ioffe2015batch}). 

\textbf{Multi-hyperparameter optimization} is not implemented in this work. We posit that this may require extending the TRP module connections from one-dimensional chains to higher-dimensional meshs, which is our current priority for exploration. 

\textbf{Neural Architecture Search} is an ongoing direction in our research. This work only demonstrates dynamic connection patterns in ResNet architectures, leaving other architectures unexplored. Notably, our approach diverges from mainstream reinforcement-learning-based methods by eliminating the gating network mechanism, which warrants future analysis of its trade-offs.

\section{Conclusion}
In this paper, we introduce Sequential Policy Gradient (SPG) modeling, a policy optimization algorithm for neural architecture search and hyperparameter optimization. We pioneer this paradigm for fast episode generation and design a compatible reward function. SPG extends the base model to a parallel input-to-padded-episode architecture by connecting with multiple temporary TRP modules, where TRP modules will be removed after training to restore the base architecture. 

The proposed approach introduces three key innovations:
(1) It supersedes conventional autoregressive trajectory collection in reinforcement learning, substantially streamlining the training process;
(2) This method adaptively optimizes hyperparameters during training, where SPG-based retraining consistently enhances base model performance. Furthermore, SPG-driven fine-tuning demonstrates superior transfer learning capabilities compared to the vanilla fine-tuning method;
(3) Extensive evaluations are performed on large-scale datasets and widely adopted models, demonstrating robust performance, cost-effective training, and high industrial applicability.

\newpage

\bibliographystyle{plainnat}
\bibliography{references}

\end{document}